{}

\documentclass{article}
\usepackage{ijcai26}

\usepackage{times}
\usepackage{soul}
\usepackage{url}
\usepackage[hidelinks]{hyperref}
\usepackage[utf8]{inputenc}
\usepackage[small]{caption}
\usepackage{graphicx}
\usepackage{amsmath}
\usepackage{amsthm}
\usepackage{booktabs}
\usepackage{algorithm}
\usepackage{algorithmic}
\usepackage[switch]{lineno}
\usepackage{booktabs}   
\usepackage{multirow}   
\usepackage{graphicx}   
\usepackage{amsmath}    
\usepackage[table]{xcolor}
\usepackage{booktabs, multirow, graphicx}
\definecolor{grayrow}{gray}{0.95}
\usepackage{subcaption}
\usepackage{booktabs}
\usepackage[table,xcdraw]{xcolor} 
\usepackage{graphicx}
\usepackage{subcaption}
\usepackage{amsmath, amssymb}
\usepackage{booktabs}
\usepackage{xcolor}

\title{Halt the Hallucination: Decoupling Signal and Semantic OOD Detection Based on Cascaded Early Rejection}

\author{
Ningkang Peng$^1$\and
Chuanjie Cheng$^1$\and
Jingyang Mao$^1$\and
Xiaoqian Peng$^2$\and
Feng Xing$^1$\and\\
Bo Zhang$^{1,*}$\and
Chao Tan$^{1,*}$\and
Zhichao Zheng$^{1,*}$\and
Peiheng Li$^{1,*}$\and
Yanhui Gu$^{1}$\thanks{Corresponding authors.}
\\
\affiliations
$^1$School of Computer and Electronic Information, Nanjing Normal University, China\\
$^2$School of Artificial Intelligence and Information Technology, Nanjing University of Chinese Medicine, China\\
\emails
\emails
zhangbo@nnu.edu.cn,
\{tutu\_tanchao, zheng\_zhichaox\}@163.com,
leees@nnu.edu.cn,
gu@njnu.edu.cn,
}

\begin{document}

\maketitle
\begin{abstract}
Efficient and robust Out-of-Distribution (OOD) detection is paramount for safety-critical applications.However, existing methods still execute full-scale inference on low-level statistical noise. This computational mismatch not only incurs resource waste but also induces semantic hallucination, where deep networks forcefully interpret physical anomalies as high-confidence semantic features.To address this, we propose the \textbf{Cascaded Early Rejection (CER)} framework, which realizes hierarchical filtering for anomaly detection via a coarse-to-fine logic.CER comprises two core modules: 1) \textbf{Structural Energy Sieve (SES)}, which establishes a non-parametric barrier at the network entry using the Laplacian operator to efficiently intercept physical signal anomalies; and 2) the \textbf{Semantically-aware Hyperspherical Energy (SHE) detector}, which decouples feature magnitude from direction in intermediate layers to identify fine-grained semantic deviations. Experimental results demonstrate that CER not only reduces computational overhead by 32\% but also achieves a significant performance leap on the CIFAR-100 benchmark:the average FPR95 drastically decreases from 33.58\% to 22.84\%, and AUROC improves to 93.97\%. Crucially, in real-world scenarios simulating sensor failures, CER exhibits performance far exceeding state-of-the-art methods. As a universal plugin, CER can be seamlessly integrated into various SOTA models to provide performance gains.
\end{abstract}

\begin{figure}[t]
    \centering
    \includegraphics[width=1.0\columnwidth]{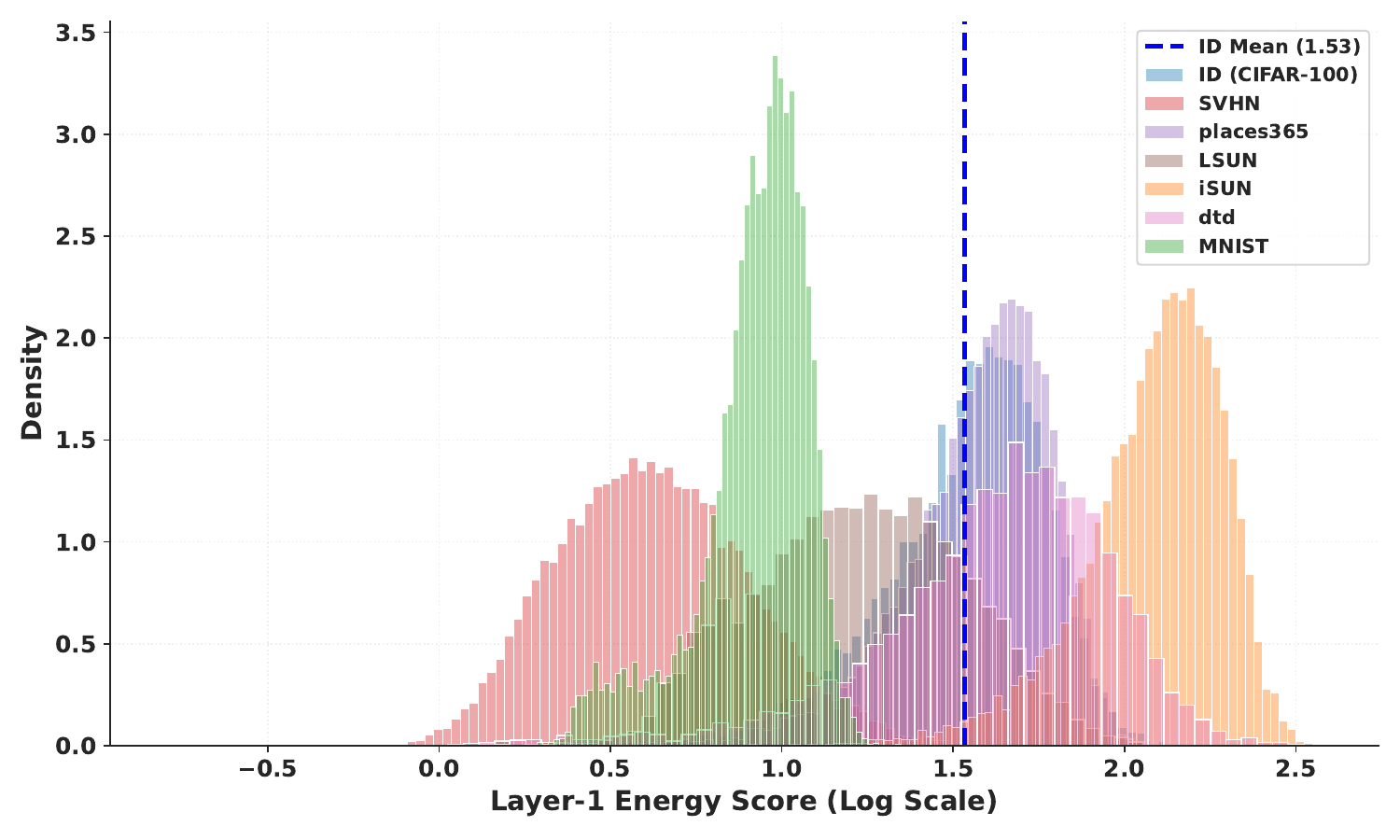}
  
    \vspace{-0.5em}
    
    \caption{\textbf{High-Frequency Energy Distribution Comparison.} The histogram illustrates the distinct separation between ID data (CIFAR-100) and various OOD datasets. Far-OOD samples like SVHN and MNIST exhibit significantly lower energy.}
    \label{fig:energy_dist}
    
    \vspace{-1.0em}
\end{figure}
\begin{figure*}[t] 
    \centering
    \includegraphics[width=0.9\textwidth]{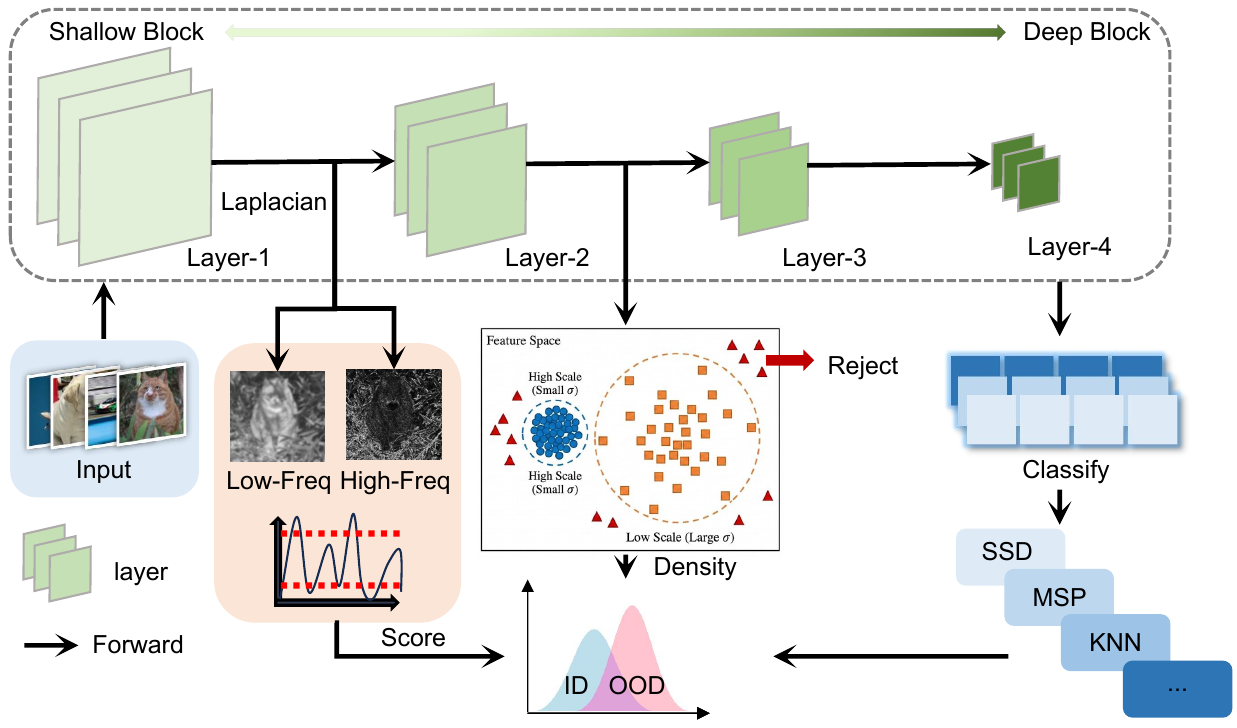} 
    \caption{Overview of proposed CER framework,framework.The \textbf{CER} framework implements an adaptive, multi-stage OOD detection mechanism by integrating physical representation analysis with deep semantic discrimination. Rather than relying on a single judgment at the final layer of the network, the framework deploys multiple specialized detection heads at varying network depths to progressively reject OOD samples based on their individual complexity.}
    \label{fig:framework}
\end{figure*}

\section{Introduction}

Deep learning models have achieved remarkable success in safety-critical domains such as autonomous driving~\cite{huang2021mos} and medical diagnosis. However, these models are typically trained under a closed-world assumption and often yield overconfident, erroneous predictions when encountering unknown Out-of-Distribution (OOD) samples in real-world open scenarios. While existing OOD detection methods have made significant strides in performance, most are confined to a post-hoc processing paradigm at the network's final layer.

This paradigm faces a severe \textbf{Computational Mismatch} in practical deployment: whether the input is a complex semantic object or simple sensor noise or physical perturbation, the model is compelled to execute full-path deep feature abstraction.\cite{Kaya2019shallow} This computational mismatch not only causes severe resource waste but, more critically, induces \textbf{Semantic Hallucination}. When non-semantic statistical noise is forcefully fed into deep networks, complex non-linear transformations over-interpret these meaningless signals, generating high-confidence misjudgments.\textit{For instance, in our preliminary experiments, a model pre-trained on CIFAR-100 would mistake simple MNIST handwritten digits for high-confidence semantic categories}. \cite{Teerapittayanon2016branchynet}This demonstrates that deep feature extraction is not only redundant but dangerous for low-level noise.

To alleviate computational bottlenecks, multi-exit architectures (e.g., MOOD) attempt to utilize intermediate layer confidence for early exiting. However, such methods often fall into the trap of \textbf{Inductive Bias Mismatch}. They forcibly assume that shallow features can simultaneously support ID classification and OOD detection. However, shallow layers, lacking global context, cannot make rigorous semantic decisions, leading models to sacrifice detection rigor for speed.

To this end, we propose the \textbf{Cascaded Early Rejection (CER)} framework. We reformulate efficient OOD detection as a cascaded inference task under computational constraints and advocate for thoroughly decoupling signal rejection from semantic classification. Inspired by the spectral energy distribution laws of natural images\cite{Wang2020high}, as shown in Figure~\ref{fig:energy_dist}, we utilize the Laplacian operator at the network entry to capture high-frequency structural anomalies, blocking deep hallucinations induced by non-semantic noise at the source. For samples that bypass the physical sieve, we utilize hyperspherical geometric projection at the semantic transition layer to decouple feature magnitude from direction,\cite{Hasnat2017von} eliminating interference from unstable activation scales and discriminating solely based on semantic directional consistency.

The main contributions of this paper are as follows:

\begin{enumerate}
    \item \textbf{Paradigm Shift:} We reveal the overlooked issues of computational mismatch and inductive bias mismatch in existing OOD detection and propose a new paradigm of decoupling shallow signal rejection from deep semantic discrimination.
    
    \item \textbf{Technical Innovation:} We introduce the synergistic SES module and SHE model. The former rapidly filters texture artifacts via physical statistical laws, while the latter ensures the semantic rigor of intermediate layer decisions via hyperspherical projection. While reducing computational overhead by 32\%, we significantly optimize the FPR95 on CIFAR-100 from 33.58\% to 22.84\%.
    
    \item \textbf{Safety Enhancement in Extreme Environments:} We conducted in-depth testing on real-world noise environments such as \textbf{Sensor Failure}. Results show that CER demonstrates robustness far exceeding existing SOTA methods, effectively preventing system crashes caused by low-level signal anomalies.
    
    \item \textbf{Highly Universal Plugin Design:} CER can be seamlessly integrated into various mainstream backbone networks, endowing models with efficient and safe detection capabilities without the need for complex retraining.
\end{enumerate}

\section{Related Work}

Out-of-Distribution Detection.Existing OOD detection methodologies generally fall into two primary categories: score-based and distance-based approaches.Score-based methods focus on designing specific metrics to discriminate between ID and OOD samples. Hendrycks \& Gimpel established a baseline with MSP, directly utilizing the Maximum Softmax Probability as a confidence score.Subsequently, ODIN ~\cite{Liang2018enhancing} introduced temperature scaling and input preprocessing to amplify the distributional divergence between ID and OOD data. ~\cite{Liu2020energy} demonstrated that the Energy Score theoretically aligns better with the probability density than softmax confidence.Furthermore, methods like ReAct ~\cite{Sun2021react} and ASH ~\cite{djurisic2022ash} suppress anomalous activations from OOD samples by rectifying or pruning feature activations.Distance-based methods, conversely, leverage geometric statistics within the deep feature space.A quintessential example is the Mahalanobis distance~\cite{Lee2018simple},which models ID features as multivariate Gaussian distributions and measures the distance between test samples and class centroids.Additionally, non-parametric approaches based on KNN have proven effective in handling complex distributions.However, existing post-hoc paradigms universally suffer from a computational mismatch.Due to their strict reliance on deep semantic features, even simple physical noise is compelled to undergo expensive, full-network inference.This not only results in computational waste but also forces the network to over-interpret low-level anomalies~\cite{Chen2025decoupling}, thereby inducing semantic hallucination~\cite{peng2025multi}.

Dynamic Inference \& Early Exiting. To alleviate the computational burden of deep models, dynamic inference architectures have been extensively studied. Classic methods such as BranchyNet~\cite{Teerapittayanon2016branchynet} and MSDNet~\cite{Huang2017multiscale}  insert auxiliary shallow classifiers at intermediate layers, allowing easy samples to exit early once a confidence threshold is met.~\cite{Zeiler2014visualizing} MOOD\cite{Lin2021mood} further extended this paradigm to OOD detection, ~\cite{Nguyen2015deep}utilizing intermediate classification confidence to intercept out-of-distribution samples early.~\cite{Scardapane2020differentiable}However, such methods suffer from an inductive bias mismatch. Shallow features inherently lack the semantic depth required for fine-grained classification; forcing intermediate layers to perform classification predictions often leads to unreliable decisions, essentially sacrificing rigor for speed.~\cite{Lid2023rethinking}In contrast, our proposed CER framework redefines the objective of early exiting.Rather than forcing intermediate layers to predict classes, we reduce the task to anomaly rejection.

\section{Method}

\subsection{The CER Framework}

\label{sec:problem_formulation}

Motivated by the efficiency of multi-exit architectures, we propose the \textbf{Cascaded Early Rejection (CER)} framework ~\cite{Bernhard2021impact}.We formulate the model as a sequential cascade of $K$ rejection modules $\mathcal{M} = \{M_1, \dots, M_K\}$. Rather than generating parallel predictions, this framework produces a chain of conditional gating decisions.

Let $\mathbf{z}_0$ denote the raw input image $x$. The feature extraction at the $i$-th stage is recursively defined as:
\begin{equation}
    \mathbf{z}_i = f_{i}(\mathbf{z}_{i-1}; \theta_i), \quad i \in \{1, \dots, K\},
    \label{eq:feature_extraction}
\end{equation}
where $f_i(\cdot; \theta_i)$ represents the $i$-th neural block parameterized by $\theta_i$.

Within the CER framework, we introduce intermediate rejection modules at varying levels to enable dynamic, coarse-to-fine OOD inference.We formally define the Rejection Gate $G_i(\mathbf{z}_i)$ at stage $i$ as a binary indicator:
\begin{equation}
    G_i(\mathbf{z}_i) =
    \begin{cases}
    \text{ 1}, & \text{if } S_i(\mathbf{z}_i) \in \mathcal{A}_i \\
    \text{ 0}, & \text{if } S_i(\mathbf{z}_i) \notin \mathcal{A}_i
    \end{cases}
    \label{eq:rejection_gate}
\end{equation}
Here, $S_i(\cdot)$ is the stage-specific scoring function, and $\mathcal{A}_i$ denotes the ID-derived acceptance region. 

Crucially, this training-free design formalizes the global inference as a conditional execution chain:
\begin{equation}
    \text{Output}(x) =
    \begin{cases}
    \perp , & \text{if } \exists i < K, G_i(\mathbf{z}_i) = 0 \\
    f_{\text{K}}(\mathbf{z}_K), & \text{otherwise}
    \end{cases}
    \label{eq:execution_chain}
\end{equation}
where $\perp$ denotes the rejection state. This formulation implies that the final semantic classifier $f_{\text{K}}$ is triggered \textit{if and only if} the sample successfully traverses all $K-1$ rejection modules.~\cite{Wolczyk2021zero}

\subsection{Physical-aware Structural Energy Screening }
\label{sec:ses}

Deep neural networks often exhibit a texture bias, assigning disproportionately high confidence to statistical anomalies. While natural images typically adhere to the $1/f^\alpha$ spectral laws, OOD samples (e.g., chaotic noise) deviate significantly from these regularities. To efficiently intercept such low-level physical anomalies without the computational overhead of the Fourier Transform, we introduce the Structural Energy Screening (SES) module.

We leverage the Laplacian operator as a computationally efficient, isotropic proxy for high-frequency spectral energy. Given the shallow feature map $\mathbf{z}_1 \in \mathbb{R}^{C \times H \times W}$,we perform depth-wise convolution with a fixed Laplacian kernel $\mathbf{K}_{Lap}$ to extract channel-wise residuals:
\begin{equation}
    \mathbf{H}_{freq} = |\mathbf{z}_1 * \mathbf{K}_{Lap}|, \quad 
    \label{eq:laplacian}
\end{equation}
where $\mathbf{H}_{freq}$ quantifies the local spectral singularities, and $|\cdot|$ denotes the element-wise absolute value. In the frequency domain, the Laplacian response approximates the integrated power spectrum weighted by spatial frequency $\omega$. Formally, the captured energy satisfies:
\[
    \mathbb{E}[|\mathbf{H}_{freq}|^2] \propto \int \omega^2 |\mathcal{F}(\mathbf{z}_1)(\omega)|^2 d\omega.
\]
This analytical result confirms that SES acts as a high-pass filter, selectively targeting the high-frequency violations typical of OOD noise.

While $\mathbf{H}_{freq}$ captures local singularities, aggregating these responses requires careful design. Standard Global Average Pooling (GAP) treats all channels equally, inevitably diluting local physical anomalies with background signals. 

To address this, we propose an \textbf{Adaptive Top-K Spectral Pooling} strategy. We focus on the most energetic frequency bands rather than the global average. The final structural energy score $S_1(\mathbf{z}_1)$ is formulated as:

\begin{equation}
\begin{aligned} 
S_1(z_1) &= \frac{1}{K} \sum_{c \in \Omega_K} \log \left( \frac{1}{HW} \sum_{h,w} H_{freq}^{(c)}(h, w) + \epsilon \right) \\ 
&= - \langle E(z_1; c) \rangle_{c \in \Omega_K} \\ 
&= \log \Psi(z_1; \Omega_K) 
\end{aligned} \label{eq:ses_score}
\end{equation}

where $H$ and $W$ denote the height and width of the feature map, respectively. The operator $\langle \cdot \rangle_{c \in \Omega_K}$ denotes the \textit{ensemble average} over the top-$K$ most energetic channels, and $\Psi(z_1; \Omega_K)$ is defined as the \textit{global confidence level} of the physical sieve. Specifically, $E(z_1; c)$ represents the channel-wise energy function derived from the spatially averaged high-frequency response, while the log-transformation $\log \Psi(\cdot)$ effectively linearizes the spectral power to suppress stochastic background noise.

Our derivation stems from the \textit{Spectral Sparsity Prior}: physical anomalies typically manifest in only a sparse subset of channels $\mathcal{C}_{anom}$ (i.e., $|\mathcal{C}_{anom}| \ll C$). 
Under the standard GAP, the anomaly signal is inevitably diluted by the dominant background responses. 
To quantify the degree of anomaly signal enhancement, we formally define the \textbf{Spectral Contrast Gain} ($G$) as:

\begin{equation}
    G \triangleq \frac{\exp(S_{1}(z_{1}))}{\frac{1}{C} \sum_{c=1}^{C} \exp(e^{(c)})}
    \label{eq:G_definition}
\end{equation}

where $S_1(z_1)$ denotes the activation strength of the Top-1 channel, and $e^{(c)}$ represents the energy response of the $c$-th channel. 
The physical significance of $G$ lies in its ability to directly characterize the ratio between the strongest single-channel anomaly activation and the ensemble average of the background noise.

Through this formulation, our SES can explicitly extract sparse anomalous features from the background. 
Consequently, even in the presence of significant texture bias, the anomaly signal remains distinguishable and is prevented from being submerged by the global averaging operation, provided that $G$ is sufficiently large.

\subsection{Semantically-aware Hyperspherical Energy }
\label{sec:she}

Inspired by related work on multi-exit architectures, we propose an energy-based OOD detection method. Specifically~\cite{Wang2025geometric}, we derive the fundamental energy score based on the concept of exit classifiers:
\begin{equation}
    E(z) = - \log \sum_{j=1}^{C} e^{f^{(j)}(z, \theta)}
\end{equation}
where $C$ is the number of classes for in-distribution (ID) data, $f^{(j)}(z, \theta)$ denotes the logit output for the $j$-th class, and $\theta$ represents the model parameters. 

In general analysis, if we consider the dominant role of feature magnitude, the energy function can be expanded as:
\begin{align}
    E(z) &= -\log \sum_{c=1}^{C} \exp(\alpha_c \|z\| + b_c) \\
         &= -\log \left( C \cdot \exp(\alpha \|z\|) \right) \\
         &= -\alpha \|z\| - \log C
\end{align}
where we assume that the feature magnitude $\|z\|$ solely determines $f^{(c)}(z)$, and the weight gains $\alpha$ are identical across all classes with zero bias, making both $\alpha$ and $C$ constants.

However, at intermediate layers of a neural network, relying solely on or even including magnitude information for decision-making is unreliable.Intermediate layers exhibit extremely strong responses to low-level physical attributes such as local contrast, sharp edges, and high-frequency noise. These physical features trigger massive feature magnitudes $\|z\|$, yet they carry no semantic information useful for distinguishing ID from OOD data. 

According to the derivation in Eq.~(10), as long as $\|z\|$ is sufficiently large, the energy will be extremely low. This creates a paradox: an OOD sample filled with meaningless high-frequency noise appears more normal to the energy model than an ID sample with a smaller feature magnitude.

To address this issue, we introduce the concept of class prototypes and hyperspherical embedding into the intermediate energy score. First, we define the class prototype $\mu_k$ as:
\begin{equation} 
    \mu_k = \frac{\sum_{n=1}^{N_k} \omega_{n,k} \cdot z_{n,k}}{\left\| \sum_{n=1}^{N_k} \omega_{n,k} \cdot z_{n,k} \right\|_2} 
\end{equation}
where $z_{n,k}$ is the feature vector of the $n$-th training sample belonging to the $k$-th class extracted from an intermediate layer, $N_k$ is the total number of samples in that class, and $\omega_{n,k}$ is the weight coefficient for the $n$-th sample, used to measure its semantic representativeness.

Next, by leveraging hyperspherical embedding~\cite{Ming2023how}, we transform the energy score into a metric focused on semantic information:
\begin{align}
    f^{(j)}(z, \theta) &= \|w_j\| \|z\| \cos(\theta_{z, w_j}) + b_j \\
                        &= \kappa_j \cdot \frac{z^\top \mu_j}{\|z\|}
\end{align}
where $\cos(\theta_{z, w_j})$ represents the semantic consistency (direction) between the feature and the class, and $\kappa_j$ denotes the intra-class angular dispersion calibration factor. ~\cite{Smith2025geometric}Based on this, we arrive at the final \textbf{semantically-aware hyperspherical energy} formula:
\begin{equation}
    E(z) = - \log \sum_{j=1}^{C} \exp \left( \kappa_j \cdot \frac{z^\top \mu_j}{\|z\|} \right)
\end{equation}

By introducing the $L_2$ normalization constraint (the $\|z\|$ in the denominator), this formula achieves complete decoupling of feature intensity from directional semantics. It forces the model to make judgments based solely on the semantic matching between the feature and the class prototypes, thereby fundamentally suppressing semantic hallucinations caused by magnitude bias.

\begin{table*}[t]
\centering

\label{table1}
\renewcommand{\arraystretch}{1.2} 
\resizebox{\textwidth}{!}{%
\begin{tabular}{lcccccccccccccc} 
\toprule
\multirow{3}{*}{\textbf{Methods}} & \multicolumn{12}{c}{\textbf{OOD Datasets}} & \multicolumn{2}{c}{\multirow{3}{*}{\textbf{Average}}} \\ 

 & \multicolumn{2}{c}{SVHN} & \multicolumn{2}{c}{MNIST} & \multicolumn{2}{c}{Places365} & \multicolumn{2}{c}{LSUN} & \multicolumn{2}{c}{iSUN} & \multicolumn{2}{c}{Textures} & \multicolumn{2}{c}{} \\ 

\cmidrule(lr){2-15} 

 & FPR95$\downarrow$ & AUROC$\uparrow$ & FPR95$\downarrow$ & AUROC$\uparrow$ & FPR95$\downarrow$ & AUROC$\uparrow$ & FPR95$\downarrow$ & AUROC$\uparrow$ & FPR95$\downarrow$ & AUROC$\uparrow$ & FPR95$\downarrow$ & AUROC$\uparrow$ & FPR95$\downarrow$ & AUROC$\uparrow$ \\ \midrule

MSP         & 78.89 & 79.80 & 88.23 & 78.25 & 84.38 & 74.21 & 83.47 & 75.28 & 84.61 & 74.51 & 86.51 & 72.53 & 84.35 & 75.76 \\
ODIN        & 70.16 & 84.88 & 92.33 & 71.54 & 82.16 & 75.19 & 76.36 & 80.10 & 79.54 & 79.16 & 85.28 & 75.23 & 80.97 & 77.68 \\
Energy      & 66.91 & 85.25 & 91.49 & 74.59 & 81.41 & 76.37 & 59.77 & 86.69 & 66.52 & 84.49 & 79.01 & 79.96 & 74.18 & 81.22 \\
Mahalanobis & 87.09 & 80.62 & 70.78 & 83.79 & 84.63 & 73.89 & 84.15 & 79.43 & 83.18 & 78.83 & 61.72 & 84.87 & 78.59 & 80.24 \\
kNN+         & 39.23 & 92.78 & 81.34 & 76.16 & 80.74 & 77.58 & 48.99 & 89.30 & 74.99 & 82.69 & 57.15 & 88.35 & 62.88 & 84.48 \\
LogitNorm    & 59.60 & 90.74 & 86.73 & 76.33 & 80.25 & 78.58 & 81.07 & 82.99 & 84.19 & 80.77 & 86.64 & 75.60 & 79.75 & 80.84 \\

CIDER       & 23.09 & 95.16 & 53.33 & 88.11 & 79.63 & 73.43 & 16.16 & 96.33 & 71.68  & 82.98 & 43.87 & 90.42 & 46.89 & 87.67 \\
PALM        & 3.03 & 99.23 & 34.46 & 93.74 & 67.80 & 82.62 & 10.58 & 97.70 & 41.56 & 91.36 & 44.06 & 91.43 & 33.58 & 92.68 \\

\midrule
\rowcolor{gray!10} \textbf{Ours} & \textbf{2.90} & \textbf{98.86} & \textbf{0.05} & \textbf{97.98} & \textbf{70.55} & \textbf{81.34} & \textbf{12.42} & \textbf{97.06} & \textbf{24.63} & \textbf{94.53} & \textbf{26.49} & \textbf{94.07} & \textbf{22.84} & \textbf{93.97} \\ \bottomrule
\end{tabular}%
}
\caption{OOD detection performance of various methods using a ResNet-34 backbone, trained on labeled CIFAR-100 as the in-distribution (ID) dataset. $\downarrow$ ($\uparrow$) indicates that smaller (larger) values are better. \textbf{Bold} numbers indicate the best results.}
\end{table*}

\begin{table*}[t]
\centering

\label{table2}
\renewcommand{\arraystretch}{1.2} 
\resizebox{\textwidth}{!}{%
\begin{tabular}{lcccccccccccccc} 
\toprule
\multirow{3}{*}{\textbf{Methods}} & \multicolumn{12}{c}{\textbf{OOD Datasets}} & \multicolumn{2}{c}{\multirow{3}{*}{\textbf{Average}}} \\ 

 & \multicolumn{2}{c}{SVHN} & \multicolumn{2}{c}{MNIST} & \multicolumn{2}{c}{Places365} & \multicolumn{2}{c}{LSUN} & \multicolumn{2}{c}{iSUN} & \multicolumn{2}{c}{Textures} & \multicolumn{2}{c}{} \\ 

\cmidrule(lr){2-15} 

 & FPR95$\downarrow$ & AUROC$\uparrow$ & FPR95$\downarrow$ & AUROC$\uparrow$ & FPR95$\downarrow$ & AUROC$\uparrow$ & FPR95$\downarrow$ & AUROC$\uparrow$ & FPR95$\downarrow$ & AUROC$\uparrow$ & FPR95$\downarrow$ & AUROC$\uparrow$ & FPR95$\downarrow$ & AUROC$\uparrow$ \\ \midrule

MSP         & 59.66 & 91.25 & 28.58 & 94.85 & 62.46 & 88.64 & 51.93 & 92.73 & 54.57 & 92.12 & 66.45 & 88.50 & 53.94 & 91.35\\
ODIN        & 20.93 & 95.55 & 69.01 & 65.23 & 63.04 & 86.57 & 31.92 & 94.82 & 33.17 & 94.65 & 56.40 & 86.21 & 45.74 & 87.17 \\
Energy      & 54.41 & 91.22 & 32.00 & 94.58 & 42.77 & 91.02 & 23.45 & 96.14 & 27.52 & 95.59 & 55.23 & 89.37 & 39.23 & 92.99 \\
Mahalanobis & 9.24 & 97.80 & 3.45 & 99.01 & 83.50 & 69.56 & 67.73 & 73.61 & 6.02 & 98.63 & 23.21 & 92.91 & 32.19 & 88.59 \\
kNN+         & 2.70 & 99.61 & 6.51 & 98.92 & 23.05 & 94.88 & 7.89 & 98.01 & 23.56 & 96.21 & 10.11 & 97.43 & 12.47 & 97.51 \\
LogitNorm    & 45.31 & 91.98 & 57.03 & 88.81 & 59.69 & 88.34 & 50.87 & 92.44 & 52.40 & 90.89 & 53.78 & 90.39 & 53.18 & 90.48 \\

CIDER       & 2.89 & 99.72 & 19.17 & 96.99 & 23.88 & 94.09 & 5.75 & 99.01 & 20.21  & 96.64 & 12.33 & 96.85 & 14.04 & 97.22 \\
PALM        & 0.52 & 99.89 & 2.95 & 99.39 & 29.31 & 94.09 & 1.78 & 99.37 & 27.24 & 95.67 & 16.67 & 97.19 & 13.08 & 97.60 \\

\midrule
\rowcolor{gray!10} \textbf{Ours} & \textbf{0.51} & \textbf{99.12} & \textbf{2.99} & \textbf{98.81} & \textbf{29.03} & \textbf{94.18} & \textbf{1.86} & \textbf{99.02} & \textbf{27.32} & \textbf{95.65} & \textbf{15.85} & \textbf{96.12} & \textbf{12.93} & \textbf{97.15} \\ \bottomrule
\end{tabular}%
}
\caption{OOD detection performance of various methods using a ResNet-18 backbone, trained on labeled CIFAR-10 as the in-distribution (ID) dataset. $\downarrow$ ($\uparrow$) indicates that smaller (larger) values are better. \textbf{Bold} numbers indicate the best results.}
\label{table2}
\end{table*}

\section{Experiments}

\subsection{Experimental Setup}

\noindent \textbf{Datasets.} We establish \textbf{CIFAR-10} and \textbf{CIFAR-100}~\cite{krizhevsky2009learning} as the in-distribution (ID) datasets. For out-of-distribution (OOD) evaluation, we utilize a comprehensive suite of natural image datasets, including \textbf{SVHN}~\cite{netzer2011reading}, \textbf{Places365}~\cite{zhou2017places}, \textbf{LSUN} (Crop and Resize)~\cite{yu2015lsun}, \textbf{iSUN}~\cite{xu2015turker},  \textbf{Textures}~\cite{cimpoi2014describing}, and \textbf{MNIST}~\cite{lecun1998gradient}. Consistent with standard protocols, all OOD images are resized to $32 \times 32$ to match the ID resolution.

\noindent \textbf{Evaluation Metrics.} We evaluate the CER framework and all baseline methods using three standard metrics: (1) \textbf{Average FLOPs}: The expected Floating Point Operations per sample during inference, explicitly including the overhead of rejection modules; (2) \textbf{FPR95}: The False Positive Rate of OOD data when the ID true positive rate is fixed at 95\%; and (3) \textbf{AUROC}: The Area Under the Receiver Operating Characteristic curve.

\noindent \textbf{Model Pre-training.}  We employ ResNet-18 for CIFAR-10 and ResNet-34 for CIFAR-100, following the experimental protocols established in~\cite{yang2022openood}. 

\begin{table}[t]
\centering

\label{table3}
\renewcommand{\arraystretch}{1.2}
\definecolor{grayrow}{gray}{0.9} 
\resizebox{\columnwidth}{!}{%
\begin{tabular}{lcccc}
\toprule
\textbf{Method} & \textbf{FLOPS(G)$\downarrow$} & \textbf{Savings.$\downarrow$} & \textbf{AUROC(\%) $\uparrow$} & \textbf{FPR(\%) $\downarrow$} \\ \midrule

\multicolumn{5}{l}{\textit{\textbf{CIFAR-100}}} \\ \midrule
MSP                     & 1.16 & —       & —       & —       \\
\rowcolor{grayrow} 
\textbf{+ CER} & \textbf{0.79 \textcolor{green!70!black}{+0.37}} & \textbf{31.75\%} & \textbf{+18.80} & \textbf{-39.34} \\

ODIN                    & 1.16 & —       & —       & —       \\
\rowcolor{grayrow} 
\textbf{+ CER} & \textbf{0.92 \textcolor{green!70!black}{+0.24}} & \textbf{21.02\%} & \textbf{+1.10}  & \textbf{-40.78} \\

Energy                  & 1.16 & —       & —       & —       \\
\rowcolor{grayrow} 
\textbf{+ CER} & \textbf{0.95 \textcolor{green!70!black}{+0.21}} & \textbf{17.78\%} & \textbf{+9.30}  & \textbf{-32.03} \\

kNN+                  & 1.16 & —       & —       & —       \\
\rowcolor{grayrow} 
\textbf{+ CER} & \textbf{0.94 \textcolor{green!70!black}{+0.22}} & \textbf{18.97\%} & \textbf{+3.38}  & \textbf{-22.4} \\

CIDER                   & 1.16 & —       & —       & —       \\
\rowcolor{grayrow} 
\textbf{+ CER} & \textbf{0.95 \textcolor{green!70!black}{+0.21}} & \textbf{18.36\%} & \textbf{+2.57}  & \textbf{-17.20} \\ 

PALM                    & 1.16 & —       & —       & —       \\
\rowcolor{grayrow} 
\textbf{+ CER} & \textbf{0.81 \textcolor{green!70!black}{+0.35}} & \textbf{31.50\%} & \textbf{+1.39}  & \textbf{-31.98} \\
\bottomrule
\end{tabular}}
\caption{Efficiency and Generalizability Analysis on CIFAR. We report the average computational cost (\textbf{FLOPS}), relative savings (\textbf{Savings.}), and performance gains in \textbf{AUROC} and \textbf{FPR} ($\Delta$ denotes improvement) when integrating CER.}
\end{table}

\noindent \textbf{Inference Configuration (CER).} Our method relies on sufficient statistics computed offline using the ID training set. To determine thresholds without data leakage, we randomly partition a held-out ID validation set (10\% of the training data).All experiments are conducted on a single NVIDIA L20 GPU.
\label{table3}

\subsection{Experimental Results}

CER outperforms a wide range of existing competitive methods. 
Table 1 summarizes various competitive baselines for OOD detection. 
All methods are trained on ResNet-34 using CIFAR-100, without assuming access to auxiliary outlier datasets. 
We compare our method with several recent competitive approaches, including 
MSP \cite{hendrycks2017baseline}, 
ODIN  ~\cite{Liang2018enhancing}, 
Energy ~\cite{Liu2020energy}, 
Mahalanobis ~\cite{Lee2018simple}, 
kNN+ \cite{sun2022knn}, 
LogitNorm \cite{wei2022mitigating}
CIDER \cite{ming2023cider}, 
and PALM \cite{jie2023palm}.

Table 1 shows that CER significantly elevates OOD detection performance. We observe that: (1) Unlike PALM, CER leverages high-frequency asymmetry for early rejection, reducing average FPR95 by 31.98\% (from 33.58\% to 22.84\%) and reaching 93.97\% AUROC. (2) CER demonstrates superior robustness on challenging datasets; for instance, it slashes FPR95 on iSUN by 30.01\% relative to PALM, proving that physical-level constraints offer vital structural criteria for decision-making. (3) Crucially, CER's computational efficiency does not compromise accuracy. By integrating high-frequency features with intermediate-layer energy scores, it achieves more discriminative and resource-efficient detection than existing SOTA methods.

As illustrated in Figure\ref{figure4a}, to better examine the internal mechanism of  CER framework, we visualize the exit distribution of samples for each OOD dataset. For the iSUN dataset, over 60\% of the samples are intercepted at the first exit, demonstrating that CER can effectively leverage physical representation features to achieve extremely early anomaly detection. Regarding the MNIST dataset, almost all data are successfully intercepted at the second exit. Furthermore, the orange line representing Computational Savings clearly indicates that the CER framework significantly reduces inference overhead while maintaining robust detection performance.

\begin{figure}[htbp]
    \centering
    \includegraphics[width=1.0\columnwidth]{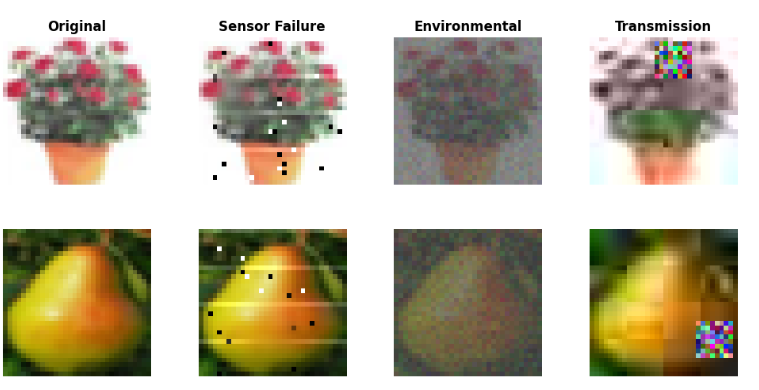} 
    \caption{Robustness analysis under extreme environments. The visualization demonstrates the distribution of detection results across different noise and failure scenarios.}
    \label{fig:robustness_plots}
\end{figure}

\begin{table}[htbp]
    \centering
    
    \label{table:robustness_results}
    \renewcommand{\arraystretch}{1.3}
    \definecolor{grayrow}{gray}{0.95}
    \resizebox{1.0\columnwidth}{!}{
        \begin{tabular}{lcccccccc}
        \toprule
        \textbf{Method} & \multicolumn{2}{c}{\textbf{Sensor Failure}} & \multicolumn{2}{c}{\textbf{Environmental}} & \multicolumn{2}{c}{\textbf{Transmission}} & \multicolumn{2}{c}{\textbf{Average}} \\
        \cmidrule{2-9} 
        & \textbf{AUC$\uparrow$} & \textbf{FPR$\downarrow$} & \textbf{AUC$\uparrow$} & \textbf{FPR$\downarrow$} & \textbf{AUC$\uparrow$} & \textbf{FPR$\downarrow$} & \textbf{AUC$\uparrow$} & \textbf{FPR$\downarrow$} \\ \midrule
        MSP             & 86.52 & 61.78 & 78.08 & 81.20 & 76.26 & 84.40 & 80.29 & 75.79 \\
        ODIN    &  67.48& 82.12 & 61.85 &88.25  & 59.50 &88.41  &  62.94&86.26  \\
        Energy    &  92.72& 40.13 & 88.89 &66.76  & 83.10 &78.45  &  88.24&61.78  \\
        kNN+     &  93.16& 41.74 & 89.07 &72.72  & 87.62 &66.88  &  89.95&60.45  \\
        CIDER    &  89.41& 70.96 & 93.03 &39.53  & 73.73 &85.42  &  85.39&66.30  \\
        PALM            & 95.57 & 25.60 & 97.87 & 8.13 & 88.47 & 61.25  & 84.30 & 48.45 \\
        \midrule
        
        \rowcolor{grayrow} 
        \textbf{Ours} & \textbf{99.22} & \textbf{3.30} & \textbf{97.29} & \textbf{5.82} & \textbf{91.50} & \textbf{28.25} & \textbf{96.01} & \textbf{12.46} \\ 
        \bottomrule
        \end{tabular}
    }
    \caption{Quantitative robustness analysis under extreme environments. We report AUROC (\%) $\uparrow$ and FPR95 (\%) $\downarrow$ for each scenario.}
\end{table}

\begin{figure*}[t]
    \centering
    \begin{subfigure}[b]{0.68\textwidth}
        \centering
        \includegraphics[width=\textwidth]{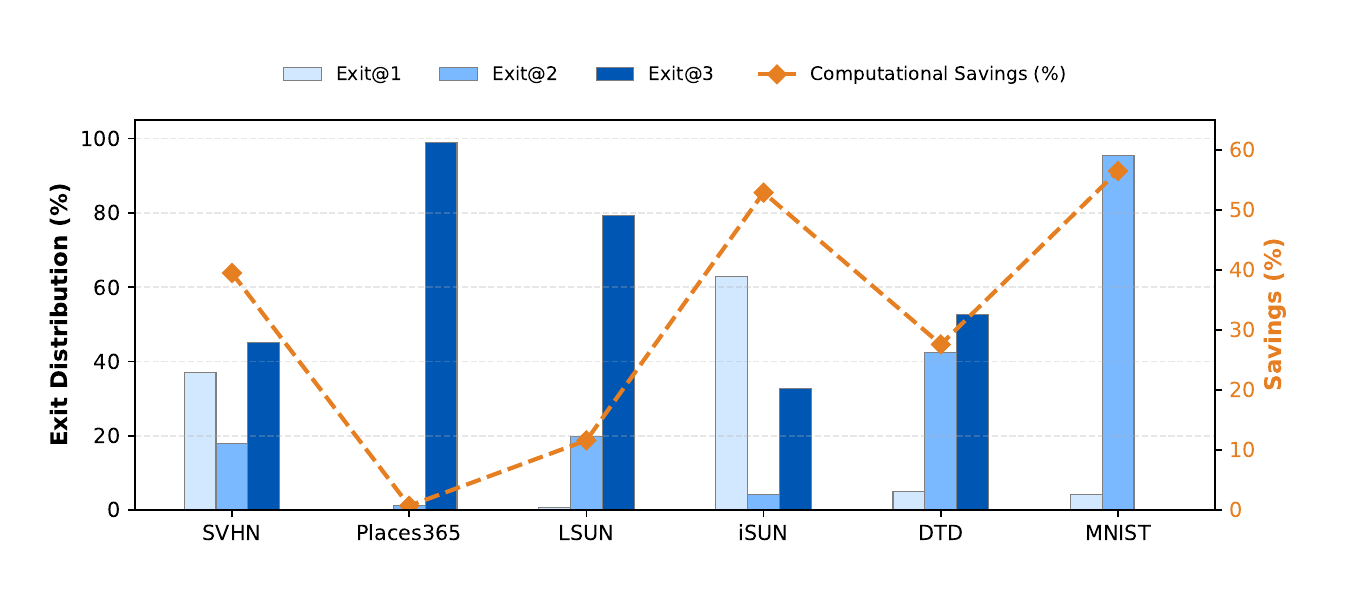}
        \caption{Exit distribution}
        \label{figure4a}
    \end{subfigure}
    \hfill
    \begin{subfigure}[b]{0.31\textwidth}
        \centering
        \includegraphics[width=\textwidth]{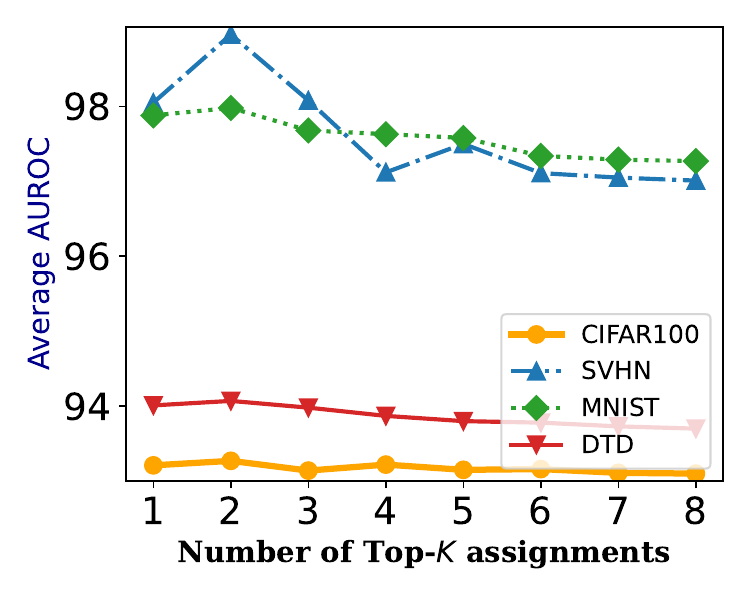}
        \caption{Top-K channel}
        \label{figure4b}
    \end{subfigure}
    
    \caption{Layer-wise energy analysis for OOD detection. (a) shows the physical frequency filtering at Stage 1, and (b) illustrates the impact of different top-$K$ channel assignments on semantic separation.}
    \label{fig:layer_analysis}
\end{figure*}

As evidenced by the comparative data in Table \ref{table2}, our proposed CER framework challenges the conventional paradigm that computational efficiency inevitably compromises detection accuracy. Specifically, CER reduces the average \textbf{FPR95 to 12.93\%}, effectively surpassing SOTA methods including PALM and CIDER on this critical metric. Despite the introduction of a demand-driven early-exit mechanism, the average \textbf{AUROC of 97.15\%} achieved by CER remains highly competitive, with a negligible performance degradation of less than 0.5\% compared to the full-path inference of PALM . Combined with the previously observed \textbf{20\%--30\%} reduction in computational cost, these results demonstrate that multi-level feature joint decision-making achieves a superior performance ceiling while maintaining exceptional resource efficiency in complex OOD scenarios.

We present CER as a versatile, plug-and-play enhancer specifically engineered for energy-efficient OOD detection. As evidenced in Table 3, integrating CER as a functional module into mainstream baselines—such as MSP, ODIN, Energy,kNN+, CIDER ,and PALM—consistently yields a dual benefit: a substantial reduction in computational overhead alongside a marked improvement in detection robustness. 

Though designed for computational efficiency, CER's multi-exit architecture supports seamless end-to-end training without exhaustive tuning. Our evaluations confirm that CER consistently yields significant computational savings, reducing FLOPs by 17.78\%--31.75\%. Crucially, these gains are coupled with a substantial boost in detection accuracy (e.g., +18.80\% AUROC for MSP) and a drastic reduction in FPR95. These results empirically validate that CER empowers models to bypass redundant feature extractions, reinforcing discriminative precision against OOD samples with a minimal computational footprint.

\begin{figure*}[t]
    \centering
    \begin{subfigure}[b]{0.23\linewidth}
        \centering
        \includegraphics[width=\linewidth]{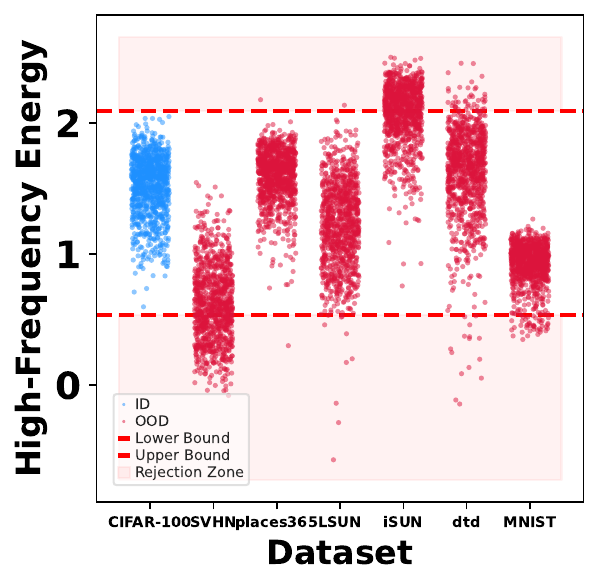}
        \caption{SES Score}
        \label{figure5a}
    \end{subfigure}
    \hfill
    \begin{subfigure}[b]{0.24\linewidth}
        \centering
        \includegraphics[width=\linewidth]{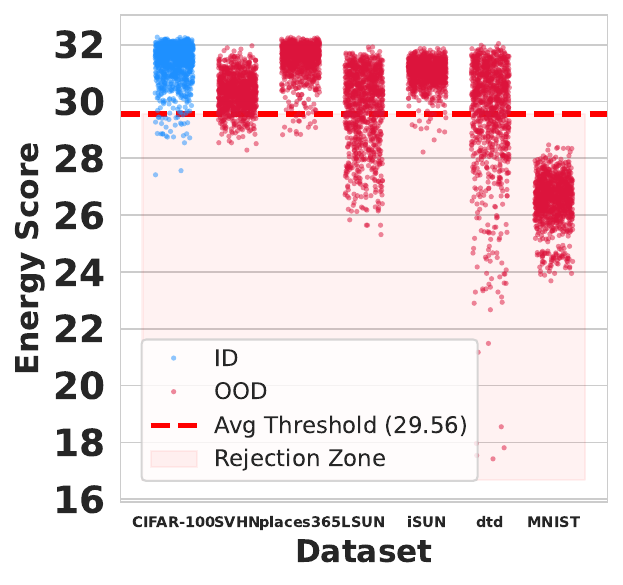}
        \caption{SHE Score }
        \label{figure5b}
    \end{subfigure}
    \hfill
    \begin{subfigure}[b]{0.24\linewidth}
        \centering
        \includegraphics[width=\linewidth]{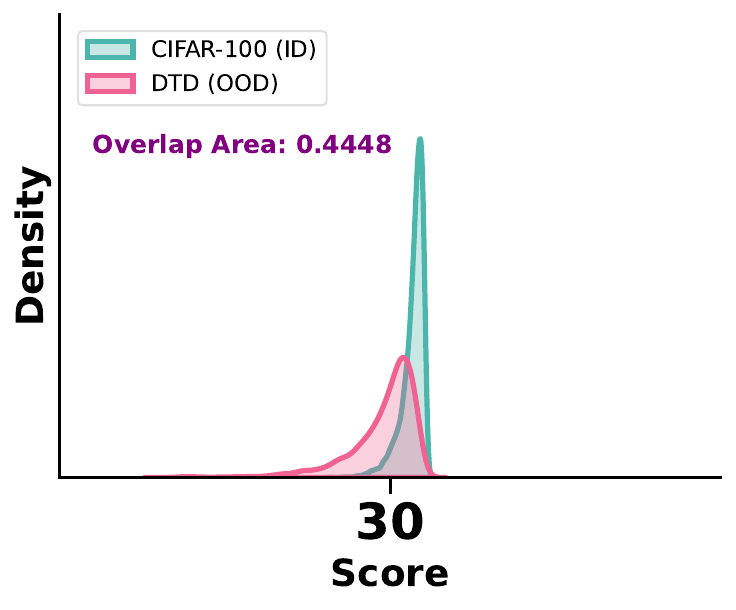}
        \caption{Normalized L2 Energy.}
        \label{figure5c}
    \end{subfigure}
    \hfill
    \begin{subfigure}[b]{0.24\linewidth}
        \centering
        \includegraphics[width=\linewidth]{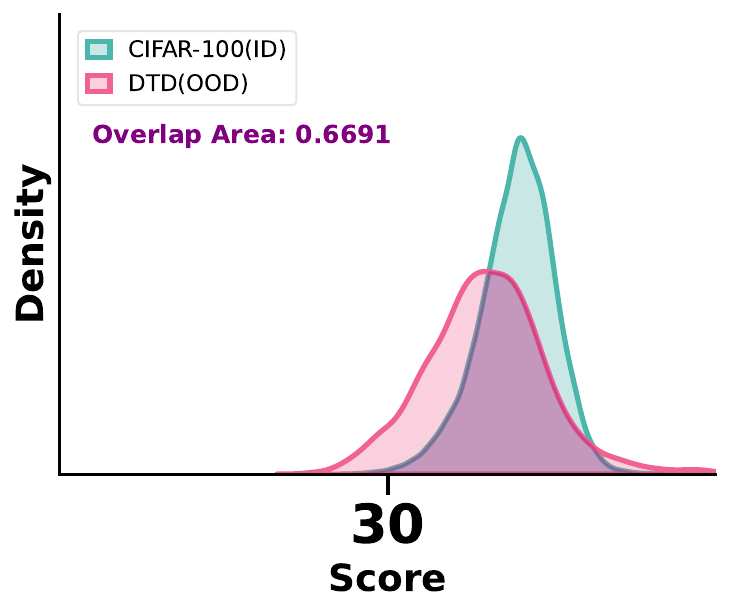}
        \caption{without L2 Normalization.}
        \label{figure5d}
    \end{subfigure}
    
    \caption{Ablation studies on different components of our method.}
    \label{fig:ablation_all}
\end{figure*}

\subsection{OOD Detection under Real-World Noise}

To evaluate the reliability of the CER algorithm in real-world deployment environments, we constructed the \textbf{CIFAR-100-R } robustness benchmark by applying simulated degradations to the complete CIFAR-100 test set . This benchmark is designed to cover a full spectrum of hardware and software failures, ranging from initial sensor acquisition to final network transmission. As illustrated in Figure \ref{fig:robustness_plots}, beyond the original images, we targeted three specific categories of corruption: sensor failure, adverse environmental interference, and transmission-compression loss.

In the Sensor Failure simulation, we focus on the most common defects in CMOS chips—dead pixels and striping noise—to evaluate the model's robustness against low-level hardware faults. Meanwhile, for Adverse Environments, we simulate visual degradation under extreme weather conditions, utilizing a composite perturbation strategy to faithfully reproduce the contrast loss and color shifts characteristic of dense fog and low-exposure environments. Furthermore, regarding Transmission Loss, we examine the impact of network communication on data quality by combining low-quality encoding with random noise patches to precisely simulate typical keyframe corruption in video streaming.

Table~\ref{table2} shows that CER outperforms all baselines under extreme stress testing on corrupted datasets. Notably, in the \textit{Sensor Failure} scenario, CER achieves a remarkably low 3.30\% FPR95, vastly surpassing Energy-based (40.13\%) and PALM (25.60\%). Even under severe \textit{Environmental} and \textit{Transmission} degradations, CER maintains robust inference by effectively leveraging residual structural information for accurate detection.

Ultimately, these stress-testing results provide strong evidence that CER possesses high practical value and robustness for real-world deployment scenarios, such as industrial monitoring and autonomous driving, where sensors are prone to diverse external interferences.

\subsection{Ablation Studies}

Selecting top-tier channels yields significant performance gains.As illustrated in Fig\ref{figure4b}, within the high-frequency energy components, we observe that the AUROC exhibits a gradual declining trend as the value of $K$ (i.e., the number of selected channels) further increases. This indicates that incorporating excessive low-energy channels introduces irrelevant physical details, which inevitably interferes with the SHE module's capability to capture semantic misalignments.

The synergy between SES and SHE underpins CER's robustness. As illustrated in Fig.\ref{figure5a}, SES acts as a first-line defense, intercepting physical anomalies (e.g., iSUN) via high-frequency energy. While SES's power is limited for semantic-shift datasets like MNIST, Fig.~5b shows that SHE fills this gap through adaptive semantic clustering. This hierarchical collaboration ensures that physical noise and semantic shifts are addressed at their respective levels, achieving exceptional performance.

Dataset-Specific Gains of L2 Normalization. Our ablation analysis reveals that L2 normalization yields substantial performance gains specifically for datasets with significant semantic shifts, such as MNIST and DTD. As illustrated in the contrast between Fig. \ref{figure5c} and Fig. \ref{figure5d}, for DTD, the application of normalization collapses the ID distribution into a compact peak, effectively reducing the overlap area from 0.67 to 0.44. This indicates that by projecting features onto a hypersphere, the SHE module can focus on angular distinguishability while ignoring magnitude-induced noise. For other datasets where the distribution shifts are less semantically distinct, the impact of L2 normalization is relatively moderate, confirming its specialized role in resolving complex semantic misalignments that bypass the initial physical-layer filtering.

\section{Conclusion}

In this study, we propose the Cascaded Early Rejection (CER) framework, a novel hierarchical inference paradigm designed to rectify the pervasive issues of computational mismatch and semantic hallucination in Out-of-Distribution (OOD) detection. CER optimizes the detection pipeline by thoroughly decoupling shallow signal rejection from deep semantic discrimination. Through the synergy of the SES and SHE modules, our framework not only significantly reduces computational overhead but also enhances overall detection performance. Notably, CER demonstrates superior early interception capabilities in safety-critical scenarios involving sensor failures and environmental noise, setting a new benchmark for reliable and energy-efficient OOD deployment at the edge.

\bibliographystyle{named}
\bibliography{ijcai26}

\end{document}